# A Methodology to Identify Cognition Gaps in Visual Recognition Applications Based on Convolutional Neural Networks

Hannes Vietz*, Tristan Rauch*, Andreas Löcklin, Nasser Jazdi and Michael Weyrich

*Abstract*— Developing consistently well performing visual recognition applications based on convolutional neural networks, e.g. for autonomous driving, is very challenging. One of the obstacles during the development is the opaqueness of their cognitive behaviour. A considerable amount of literature has been published which describes irrational behaviour of trained CNNs showcasing gaps in their cognition.

In this paper, a methodology is presented that creates worst-case images using image augmentation techniques. If the CNN's cognitive performance on such images is weak while the augmentation techniques are supposedly harmless, a potential gap in the cognition has been found. The presented worst-case image generator is using adversarial search approaches to efficiently identify the most challenging (worst) image. This is evaluated with the well-known AlexNet CNN using images depicting a typical driving scenario.

## I. INTRODUCTION

Over the last years neural network applications have increasingly been used for critical industrial operations like for instance predictive maintenance [1] and quality control in manufacturing [2, 3], with the obvious pinnacle of a high-risk application being autonomous driving. One of the important applications of neural networks is vision-based object detection with convolutional neural networks (CNN), which is often used to give the car a semantic understanding of its environment via its camera. It is e.g. critical for an autonomous vehicle to understand if it is seeing another car or a street sign. Object detectors based on the CNN technique have also been increasingly used for other industrial use cases [4].

CNNs for vision recognition applications are typically trained using large amounts of labeled images to detect specific objects in them. The inner workings, i.e. how exactly the object detector CNN arrives at its detections, is mostly incomprehensible to the human observer. Meaning, it is very hard to judge which connections between its input images and the detected objects a CNN has learned and how strong these connections are. Hence, they are often referred to as black boxes due to their opaque decision process.

An often referenced testimony of this opaqueness in the decision process is an early project of the US Army to detect camouflaged tanks amid trees [5]. The researchers trained an object detector to decide if a tank is in the image. On the first glance the results looked promising, but after a more thorough investigation it was discovered that all images containing tanks were taken on cloudy days and the images without tanks on sunny days. The detector had actually learned to decide based on the weather condition instead of detecting hidden tanks. Another akin behaviour of a CNN has been demonstrated by training a detector to distinguish dogs and wolfs [6]. Most of the wolf images had a snowy background, while the dog images were snow free. The authors showed that the trained CNN based its decisions mostly on the snow and not the animal in the image. In the following, we will call these wrongly learned hidden connections cognition gaps.

The Problem is apparent: Manually identifying cognition gaps of a CNN for visual recognition applications without any computational assistance is cumbersome: An engineer needs to look through all training and test images trying to come up with a misleading pattern, the objects might have in common. After finding a potential pattern, the engineer needs to create test images to verify his suspicion. For large datasets this manual approach is hard, costly and time consuming. Methodologies and tools are needed, that help engineers to identify cognition gaps more effectively.

The Objective of this paper is to presents a methodology to identify specific cognition gaps in CNNs, which are trained for visual recognition tasks. Main part of the methodology is a data generator that creates iteratively more difficult test images based on an initial input by an engineer. The input defines a specific change to the data, which should not change the output of the CNN. Based on the most difficult image created, i.e. the worst-case image, the engineer can decide if there is a cognition gap regarding his specific input.

The Structure of this paper is as following. First, we present an overview of current approaches that help engineers to identify cognition gaps in CNNs in Section II. More specifically we discuss approaches that visualize cognition and techniques that create test data showcasing cognition gaps. In Section III we present our proposed methodology for identifying cognition gaps in visual recognition applications based on CNNs by creating a test data generator. This generator is prototypically realized in Section IV and

Hannes Vietz, M.Sc. is scientific staff member at the Institute of Industrial Automation and Software Engineering at University of Stuttgart, Stuttgart, 70569 Germany (corresponding author, phone: +49 711-685- 67294; fax: +49 711-685-67302; e-mail: hannes.vietz@ias.uni-stuttgart.de).

Tristan Rauch, B.Sc. is a student at the Institute of Industrial Automation and Software Engineering at University of Stuttgart.

Andreas Löcklin, M.Sc. is scientific staff member at the Institute of Industrial Automation and Software Engineering at University of Stuttgart (e-mail: andreas.loecklin@ias.uni-stuttgart.de).

Dr.-Ing. Nasser Jazdi is deputy head of the Institute of Industrial Automation and Software Engineering at University of Stuttgart (e-mail: nasser.jazdi@ias.uni-stuttgart.de) and Senior Member of the IEEE.

Prof. Dr.-Ing. Dr. h. c. Michael Weyrich is head of the Institute of Industrial Automation and Software Engineering at University of Stuttgart (e-mail: michael.weyrich@ias.uni-stuttgart.de).

*These authors contributed equally to this publication.

evaluated in Section V. We conclude this work by describing potential added value for engineers, the limitations of our methodology and an outlook for further research.

## II. RELATED WORK

This section introduces related work regarding the identification of cognition gaps in CNNs. The first Part A will briefly introduce techniques that aim to visualize the cognition of CNNs, which can help to identify cognition gaps. Part B describes methodologies that try to find cognition gaps by observing the neuron activity inside CNNs. In the last Part C a special kind of cognition gap, the adversarial cognition gap, is presented and it will be elaborated how it can be found.

### A. Visualizing cognition of convolutional neural networks

The visualization of the cognitive process of CNNs is widely studied as part of the larger research field of explainable artificial intelligence. First, we focus on the visualization of CNN's cognitive ability for visual recognition on test images. One of the early works of this field studied the cognition of the aforementioned Wolf-/Dog detection CNN [6]. The researchers trained an interpretable linear model locally around their CNN's predictions on test images. Interpreting this linear model revealed which pixels of input images the CNN uses to come to its conclusion and how important they are for the decision. This can be thought of as a heat map of the CNN's cognition on a single image. By then studying several of those heatmaps, the researchers estimated that the CNN relied mostly on snow being in the image to detect wolfs. They validated these findings by giving the CNN a husky in front of a snowy landscape to predict, which was recognized as a wolf by the CNN.

It has also been demonstrated that the learned features of the inner layers of CNNs can be visualized [7]. This is achieved by giving the CNN numerous test images and observing which layers respond to specific parts of the images, resulting in a map depicting the cognition landscape of a CNN [8]. The cognition landscapes show via abstract shapes, colors and textures which features the CNN uses for its recognitions. Researchers can study these maps for guidance when they try to find cognition gaps as well as explain strange behaviour of their trained vison model.

Both presented visualization approaches are powerful techniques that can help researchers to find cognition gaps in their CNNs. However, they do not create test images that demonstrate the cognition gap and they tend to be very computationally expensive, sometimes taking several hundred hours of GPU computation time [8].

### B. Searching for cognition gaps guided by neuron activity

While the inner workings of CNNs are mostly incomprehensible to us without the earlier presented visualization techniques, researches suggested that the pure degree of neuron's activations can be used to find cognition gaps [9]. The basic assumption being, that the more neurons are active in the CNN's layers, the more error prone its recognition ability is. They created a neuron coverage metric and an optimization algorithm that aims to greedily increase the neuron coverage by applying changes to the image. The image changes like scaling, adding artificial mist, rotating and shearing have the intriguing property to not change the object, which is supposed to be recognized by the CNN. Which means they can be applied to a labeled test image without changing its correct label. Later on, in this work, we will use other image changes with the same property and call them recognition-invariant image changes. Using the neuron coverage technique, the researchers identified several cognition gaps in CNNs that earlier won challenges for their high accuracy. Unfortunately, the approach has several downsides: one being the tendency to find a lot of false positives, which meant the researches had to go through several thousand found images manually to identify the ones showing cognition gaps. Another downside is the missing of several other cognition gaps in the CNNs, as has been pointed out later by another group [10].

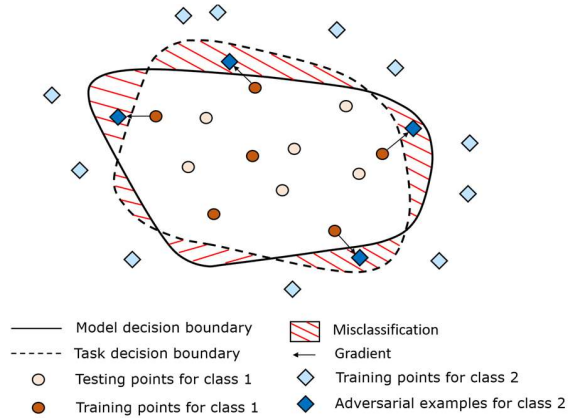

**Figure 1:** Origin of adversarial examples

### C. Adversarial cognition gaps

Adversarial Examples were first described by Szegedy [11] in 2014 and can be understood as inputs to neural networks that have been manipulated by small, targeted perturbations. The manipulations work in a way, that they cause the neural network to misclassify, despite showing no visible perturbations to a human, which he could use to identify a cognition gap.

There are now countless approaches to generating adversarial examples, but probably the best known and most popular approach, is the fast gradient sign method (FGSM) described by Goodfellow [12]. Thereby the gradient of the loss w.r.t the input data $\nabla_x J(\theta, x, y)$ is used to adjust the input data $x$ to maximize the loss.

$$x' = x + \epsilon \cdot sign[\nabla_x J(\theta, x, y)] \quad (1)$$

In (1) $x$ represents the original input data, $y$ the associated ground truth label, $\theta$ the model parameters of the neural network and $\epsilon$ a multiplier to ensure that the disturbance does not turn out too large and thus become "visible". The gradient is determined via backpropagation as when training the network. To speed up the generation of the image, only the signs of the gradient are used. The adversarial example results from the superposition of the original image with the disturbance. With a high probability, the generated image is now misclassified. One explanation for the existence of adversarial examples and probably all other cognition gaps is that the decision boundaries of the task, as seen from the human perspective, and the trained model often do not match exactly. Adversarial examples lie in the border area of the decision boundaries, as shown in **Fehler! Verweisquelle konnte nicht gefunden werden.** [12]. Adversarial examples are robust across various machine learning algorithms and neural network models [12] and can also fool classifiers in the physical world [13].

If a neural network is vulnerable to a specific adversarial attack and is to be made robust against it, defense methods must be applied. Although there are approaches adversarial training [11] there are still no comprehensive methods that protect against a broad spectrum of different adversarial attacks quickly and with adequate effort.

Adversarial search methodologies have often been successfully used to create images that showcase cognition gaps in CNNs. Additionally, the techniques are mostly very efficient. Using the earlier discussed FGSM just takes a few seconds on modern hardware, even when larger CNNs like AlexNet [14] are used [12]. But the practicality of the created images showcasing adversarial cognition gaps is often ambiguous, because they are so unspecific. What conclusions can an engineer draw from finding adversarial cognition gaps for his further work of improving his CNN? He knows that they exist and that he might need to implement one of the earlier mentioned defense mechanisms, but gets no specific information on how to improve his CNNs accuracy for its real-world recognition tasks.

*Conclusion of Related Works*

Cognition visualizations can help engineers to manually find cognition gaps by showing them parts of the inner workings of CNNs and on which parts of images the CNN focuses, but are computationally very expensive to create. Neuron activity guided techniques find test images that showcase cognition gaps, but are prone to false positives, which lead to manual work for engineers. But after the manual work of finding the images with actual cognition gaps is done, the images can be used to diagnose specific problems in the CNNs real-world recognition of objects, which is very helpful to engineers. Adversarial search algorithms are very efficient and create images showcasing cognition gaps reliably, but do not help engineers in getting an understanding of problems concerning specific recognition tasks of his CNN. Approaches that help engineers to efficiently find cognition gaps regarding specific recognition problems are missing.

### III. METHODOLOGY FOR THE TEST IMAGE GENERATOR TO REVEAL COGNITION GAPS

In the following section the proposed methodology for a worst-case test image generator, that helps an engineer to identify specific cognition gaps in CNNs is presented. First, the idea underlying the generator, recognition-invariant image changes, is presented. Then the methodology how to realize such an image generator is explained step by step.

*A. Recognition-Invariant Image Changes*

The basic idea of identifying cognition gaps is to establish an efficient way of evaluating if the CNN's engineer and the CNN itself have a similar understanding on the nature of the CNN's objective. In other words: If the decision boundaries of the engineer and the CNN align for a specific problem. An example for a CNN based object detector, that is supposed to detect cars should help with the better understanding. The statement **A**: "*The color of a car does not determine that it is a car.*" is trivial and most humans would immediately agree to it. If we found one to disagree, it seems reasonable to assume that they have a wrong understanding of what a car actually is and we would not trust them to detect cars reliably. The earlier discussed existence of adversarial test cases and the examples of the camouflaged tanks and the wolves show that CNNs often have a very different understanding of objects they detect to humans. It therefore seems to be a worthwhile endeavor to try to determine if a CNN would agree to the earlier statement **A**. If we discovered that a CNN's decision on an object being a car is heavily dependent on its color, we might deduce that there is a gap in the cognition of the CNN for its ability to detect cars. On the other hand, if it can be shown that the network is sharing the human notion on cars, we can increase our trust in its ability to detect cars. What we did by formulating statement **A**, was defining a change to the input image, which is not supposed to change the predicted output of the CNN. This special type of change is further on called recognition-invariant image change. We propose to create a test image generator, which will apply recognition-invariant image changes, with the aim of worsening the output of the CNN.

*B. Methodology to find cognition gaps*

This section presents our methodology to identify cognition gaps in CNNs by checking if recognition-invariant image changes are actually not changing the recognition of the CNN. Figure 2 shows the steps of the method and in the following paragraphs each step is explained in more detail.

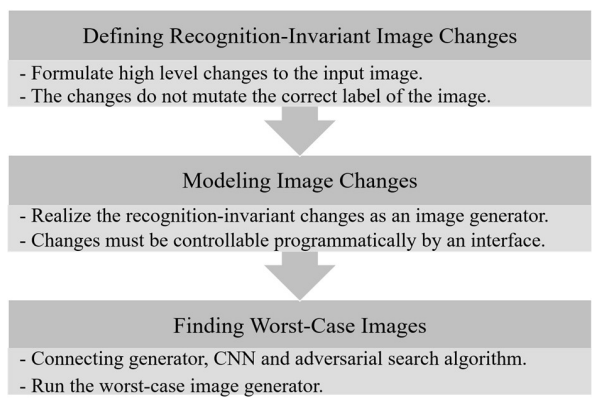

**Figure 2:** Methodology to create a worst-case image generator

*1) Defining Recognition-Invariant Image Changes*

The first step of the proposed methodology consists of defining recognition-invariant image changes, e.g. statement **A**. The changes need to be human understandable and define an abstract change in the input data, which is not supposed to change the CNN's correct recognition of objects. Changes like this are sometimes called metamorphic transformations. In contrast to earlier discussed image augmentation techniques like shearing the recognition-invariant image changes should test a more abstract understanding of the CNN.

*2) Modeling Image Changes Using an Image Generator*

In this step the earlier defined recognition-invariant image changes e.g. **A** are realized in the form of an image generator. Possible implementations are simulators that create new images from the ground up or an augmentation stack that uses existing images and changes them to create new unseen images. Simulators can create a larger variety of new images and therefore realize more possible recognition-invariant image changes than classic augmentation, but might need significantly more resources. Independent of the type of implementation the image generator needs to expose an interface which controls the location and severity of the image changes.

*3) Finding Worst-Case Images*

The last step of the proposed methodology is searching for the worst-case image, which might reveal a cognition gap to the engineer. An adversarial search methodology as described in II.C is used to iteratively find input images on which the network performs poorly. Figure 3 shows this setup, consisting of the CNN under test, the image generator and the adversarial search algorithm. The adversarial search algorithm controls the image generator with its interface and optimizes it to create challenging images for the CNN. The network's output is then processed by the adversarial search algorithm to in turn control the image generator to create even more challenging images. Setting the right number of iterations is critical. Choosing too few iterations will result in missed challenging image changes, while too many iterations will waste computational resources. It is therefore advisable to when in doubt choose a larger number. When a resource intensive image generator is used the iterative process can be stopped early, if the CNN under test's performance is not changing significantly enough. Another possibility is to keep increasing the number of iterations until the network's resulting performance is not changing anymore, which the one we apply in the next section.

*C. Output of the Worst-Case Image Generator*

Once the earlier described iterative process has ended, the input image on which the CNN performed worst is saved. By studying this image, the engineer can decide if the CNN's cognition contains gaps. If the accuracy of the object the CNN is supposed to detect has only decreased by a few percentages it is reasonable to assume, that the CNN has no cognition gap for this specific object and image change. On the other hand, if the CNN's output has changed drastically, the engineer can assume that it has learned something that is contrary to his understanding of the specific object recognition task, which suggests the existence of a cognition gap.

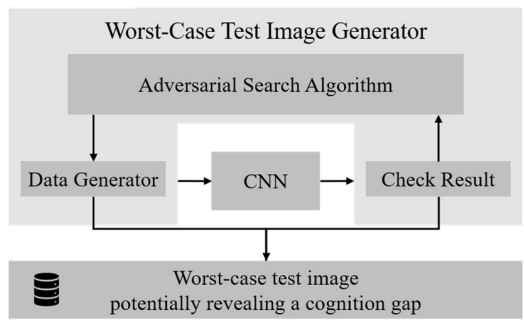

**Figure 3:** Schematic of the resulting Worst-Case Test Image Generator

IV. REALIZATION OF THE METHODOLOGY

This section describes how the proposed methodology was used to prototypically implement a worst-case image generator for the recognition of cars by the well-known object detector CNN Alex Net [14].

*B. Applying the proposed methodology*

In the following section the proposed methodology is applied step by step to realize the test *image generator*.

*1) Defining Recognition-Invariant Image Changes*

Our prototype *test image generator* shall be used on images containing cars. Some of the potential recognition-invariant image changes are:

A. *"The color and texture of a car does not determine that it is a car."*
B. *"The content of the licence plate of a car does not determine that it is a car."*
C. *"The shape of the headlights of a car does not determine that is a car."*

In the prototype developed for this study we focus on **A**, which we already discussed in the section before. The prototype shall check if the network under evaluation is relying heavily on the color and texture of a car as a feature, despite this being mostly pointless from a human perspective.

*2) Modeling Image Changes Using the Image Generator*

The data generator is implemented as an image augmentation stack realizing the by **A** defined changes. First, we chose an image of a car that is not part of the ImageNet dataset. Figure 4 contains the chosen image. Then all painted pixels of the car were segmented by hand in an image processing software. All augmentations to the image that will result in new image data are restricted to the segmented pixels. While the resulting image generator does create images that test the in **A** defined recognition-invariant image changes, they show highly unrealistic color patterns. To mitigate this, we additionally limit the allowed augmentations to the color channel red. The resulting segmentation mask and color restriction is saved and used in the next step.

*3) Finding Worst-Case Images*

The adversarial search method used is the in Section II.C described FGSM method, which we implemented ourselves.

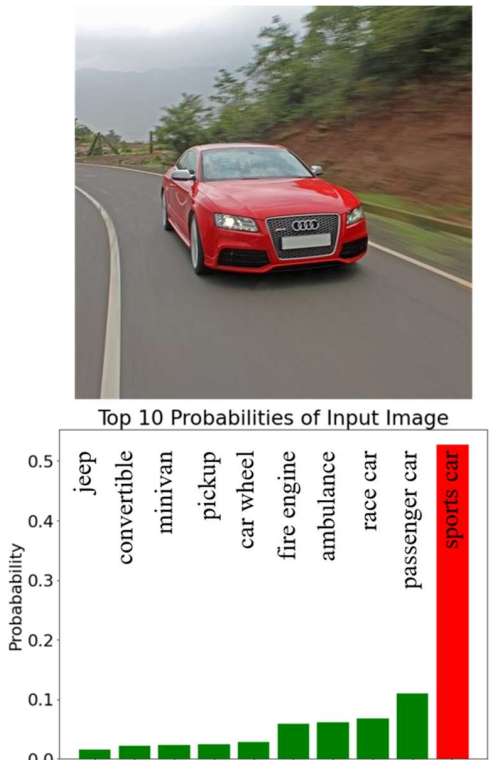

**Figure 4:** The CNN successfully detected a car in this test image. The ten highest class probabilities are all types of cars or car parts. The image is licence free and free for private and commercial usages from *pixabay.com*.

According to [12] FGSM performs better when a target class is used. Therefore, we used the class *street sign*, because of the obvious catastrophic implications following a confusion of another car with a street sign by an object detector. Meaning, FGSM will try to change the predicted class to *street sign*. It is only allowed to change the CNN's input images via the interface defined in step 2 of the methodology. By that, we limit FGSM's changes to the pixels that are in the segmentation mask and adhere to the color restriction. Our implementation of FGSM is changing the depicted car image rapidly, as can be seen by the generated images of the first six iterations in Figure 5, which means that the gradient of the CNN's loss function is very high for these changes. Iterating was stopped after a *street sign* was predicted with ~100%.

### C. Technical setup and performance of the experiments

The experiments were run on a PC running Ubuntu 20.04 LTS with a GeForce RTX 2080 GPU and an AMD Ryzen Threadripper 2920X CPU using the PyTorch deep learning framework. The total runtime of the 15 iterations is well under *1.0s* on the aforementioned machine.

## V. EVALUATION

In this section the results of the in Section IV realized worst-case image generator are presented and evaluated.

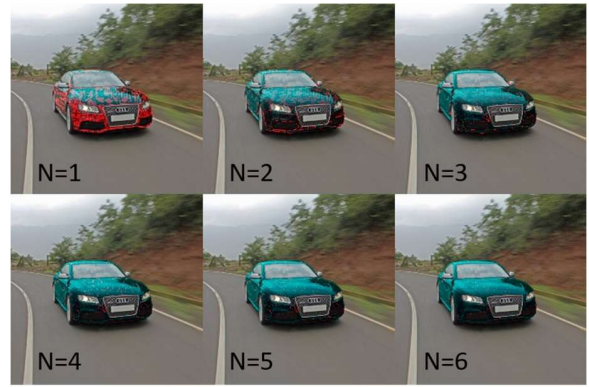

**Figure 5:** The first 6 of total 15 images created by the generator. Every car is increasingly harder to recognize correctly for the CNN.

### A. CNN's performance on the unchanged car image

First, we will discuss the CNN's performance on the initial image of a car that was chosen as a starting point of the data augmentation stack contained in the worst-case image generator. Figure 4 shows the image and the top 10 predicted probabilities of the objects in the picture. As can be seen there is no single class *car* but several more specific classes like *sports car*, *passenger car*, *jeep* and car parts like *wheel*. While some detected classes like *jeep* and *fire engine* are not perfectly fitting the pictured vehicle, they are all car related and we can therefore ascribe the CNN a very high accuracy in detecting the car, which is not surprising. AlexNet won the ImageNet Large Scale Visual Recognition Challenge in 2012.

### B. CNN's performance on the worst-case image

The last created image of the prototype worst-case image generator and the top 10 output probabilities for it can be seen in Figure 6. The CNN detects a street sign with ~100% probability. The other 9 classes, whose probabilities are near 0%, but are still in the top 10 are all completely unrelated to cars like *pole* and *fence*. These results suggest that the CNN is relying heavily on the color and texture of a car to decide if it is in fact a car. The image quality of the green car is reduced compared to the original image. Some detail is lost, especially on the edges, because the segmentation mask was created by hand and is therefore not perfect. While the restriction of the color channels on red kept some details like the mirroring of a tree and white highlights on the engine hood intact, the car still became blurry due to the pixel fluctuations caused by the augmentations. However, the results clearly show that the CNN under test is relying heavily on detailed structures of the car's color and completely ignores the, to a human, obvious car features like the shape, the lights and the grille, which suggests a cognition gap for this specific recognition task.

## VI. CONCLUSION

The presented methodology can be used by engineers to create worst-case image generators which help them to identify cognition gaps in their CNNs early in the development process. We suggest the engineer focuses on a very specific problem like the presented car example. Then the engineer creates worst-case image generators based on his understanding of the problem as outlined in Section III. The found worst-case images give the engineer hints on what his CNN has learned and if the CNN shares his understanding of the problem. The high performance of the presented approach makes it feasible to use it as a fast and low-cost tool directly after the training of a CNN, comparable to unit tests in traditional software engineering before more elaborate and costly tests in the field are being run. Even when several worst-case image generators are used, the runtime of the implemented prototype will remain in the range of seconds, unless a large simulation is used as image generator. The engineer can use the additional information provided by the generated worst-case images to improve his further training process, e.g. change his image collection strategy, adapt his loss function or start to use data augmentation techniques during future trainings.

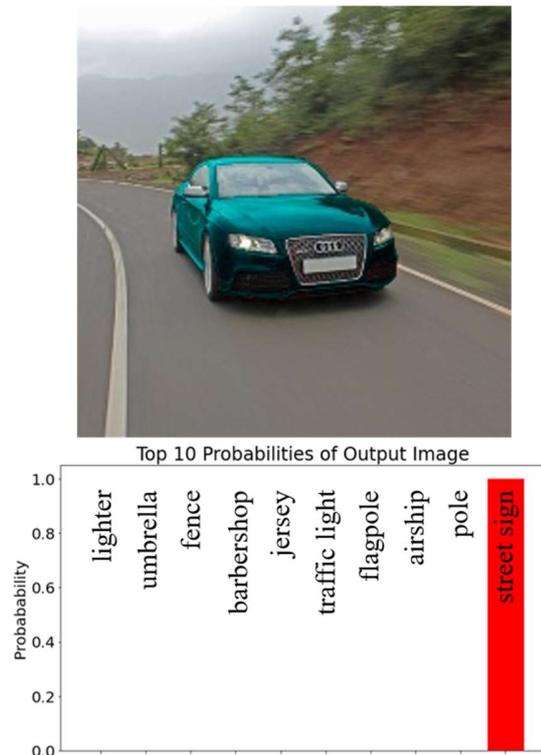

**Figure 6:** The CNN has failed to detect the car on the 15th created image and detects a street sign with ~100% certainty.

The proposed methodology is limited to checking for a similarity of understanding between the CNN and a human engineer for a very specific problem utilizing the defined recognition-invariant image changes. Checking for perfect similarity seems impossibly hard or very expensive for most practical use cases: If one could find a definite formal description of the CNN's task which seems necessary for a perfect understanding test, the usage of a CNN would be unnecessary and one could simply write a classic program to fulfill the objective. However, a more comprehensive similarity check than presented in this study seems possible, which is discussed in the following outlook on future work.

A further study could asses on how efficient a simulator can be integrated into a worst-case image generator in combination with a more advanced adversarial search algorithm. Another natural progression of this work, inspired by adversarial training, is, to use the generated images which revealed a cognition gap to retrain a CNN. It could then be further studied if the resulting CNN's cognition gaps decrease and how the accuracy changed, compared to the original test images. Additionally, the created images might be used to replace expensively collected real images and thereby increase the efficiency of the development process of CNNs.

## VII. REFERENCES

[1] B. Maschler, H. Vietz, N. Jazdi, and M. Weyrich, Eds., *Continual Learning of Fault Prediction for Turbofan Engines using Deep Learning with Elastic Weight Consolidation*, 2020.

[2] B. Lindemann, N. Jazdi, and M. Weyrich, Eds., *Anomaly detection and prediction in discrete manufacturing based on cooperative LSTM networks*, 2020.

[3] B. Lindemann, T. Müller, H. Vietz, N. Jazdi, and M. Weyrich, Eds., *A Survey on Long Short-Term Memory Networks for Time Series Prediction*, 2020.

[4] B. Maschler, S. Kamm, and M. Weyrich, "Deep Industrial Transfer Learning at Runtime for Image Recognition. at," *Automatisierungstechnik*, vol. 3, 2021.

[5] E. Yudkowsky and others, "Artificial intelligence as a positive and negative factor in global risk," *Global catastrophic risks*, vol. 1, no. 303, p. 184, 2008.

[6] M. T. Ribeiro, S. Singh, and C. Guestrin, ""Why Should I Trust You?"," in *Proceedings of the 22nd ACM SIGKDD International Conference on Knowledge Discovery and Data Mining*, San Francisco California USA, 08132016, pp. 1135–1144.

[7] C. Olah, A. Mordvintsev, and L. Schubert, "Feature Visualization," *Distill*, vol. 2, no. 11, 2017, doi: 10.23915/distill.00007.

[8] S. Carter, Z. Armstrong, L. Schubert, I. Johnson, and C. Olah, "Activation atlas," *Distill*, vol. 4, no. 3, e15, 2019.

[9] Y. Tian, K. Pei, S. Jana, and B. Ray, "DeepTest," in *Proceedings of the 40th International Conference on Software Engineering*, Gothenburg Sweden, 05272018, pp. 303–314.

[10] J. Kim, R. Feldt, and S. Yoo, "Guiding deep learning system testing using surprise adequacy," in *2019 IEEE/ACM 41st International Conference on Software Engineering (ICSE)*, 2019, pp. 1039–1049.

[11] Christian Szegedy *et al.*, "Intriguing properties of neural networks," in *International Conference on Learning Representations*, 2014.

[12] Ian Goodfellow, Jonathon Shlens, and Christian Szegedy, "Explaining and Harnessing Adversarial Examples," in *International Conference on Learning Representations*, 2015.

[13] J. Li, F. Schmidt, and Z. Kolter, "Adversarial camera stickers: A physical camera-based attack on deep learning systems," in *International Conference on Machine Learning*, 2019, pp. 3896–3904.

[14] A. Krizhevsky, I. Sutskever, and G. E. Hinton, "ImageNet Classification with Deep Convolutional Neural Networks," in *Advances in Neural Information Processing Systems*, 2012, pp. 1097–1105.